# A Grid-Point Detection Method based on U-Net for a Structured Light System

Dieuthuy Pham[1, 2], Minhtuan Ha[1, 2] and Changyan Xiao[1]

[1]College of Electrical and Information Engineering,
Hunan University, Changsha 410208, China
[2]Faculty of Electrical Engineering, Saodo University,
Haiduong 170000, Vietnam

*ABSTRACT*

*Accurate detection of the feature points of the projected pattern plays an extremely important role in one-shot 3D reconstruction systems, especially for the ones using a grid pattern. To solve this problem, this paper proposes a grid-point detection method based on U-net. A specific dataset is designed that includes the images captured with the two-shot imaging method and the ones acquired with the one-shot imaging method. Among them, the images in the first group after labeled as the ground truth images and the images captured at the same pose with the one-shot method are cut into small patches with the size of 64x64 pixels then feed to the training set. The remaining of the images in the second group is the test set. The experimental results show that our method can achieve a better detecting performance with higher accuracy in comparison with the previous methods.*

## 1. INTRODUCTION

One-shot structured light methods are being more and more developed since its advantages in terms of reconstructing the dynamic scene. Besides, these methods are also very economical and easy to do with just one camera and one projector. Firstly, a well-coded pattern is projected onto the scene. Then the image of the deformed pattern captured by the camera is fed into a decoding program to find correspondence points in the projector image plane and the camera image plane. With the points obtained, the triangulation method will be performed to find the 3D points in the point cloud. Although these methods cannot provide a dense 3D map with high accuracy as the multi-shot ones, it still satisfies the requirements of many applications in reality.

Up to now, numerous one-shot imaging methods have been proposed with different pattern coding strategies and achieved some achievements [1]. A multi-stripe pattern with De Bruijn sequence single-axis coded is proposed in [2]. With the constraint that two consecutive stripes cannot share the same color channel, so many color channels are required for a large resolution of the pattern. In this method, the pixels on the centerline of the color stripes are taken as feature points, thus the 3D maps are highly dense. However, it is sensitive to color noise and improper to reconstruct the dynamic scene. Lei et al. [3] introduced a 6-symbol M-array pattern designed with a 3 x 3 property window directly driven by the Hamming distance. The symbols are classified according to their geometrical feature and their centers are selected as feature points. With the advantage of using a binary pattern, this method can capture colorful objects and solve the discontinuity problem. Although this method is easy to decode, its drawbacks are false detection risk and sparse 3D map. To overcome the challenges of the method using the color stripe pattern, a pattern of the self-equalizing De Bruijn sequence, scale-space analysis, and band-pass complex Hilbert filters were introduced in [4]. With a pattern of color stripes separated by a gap that allows two consecutive stripes to share the same color channel, the method is capable of accurately restores the color of the stripes of the pattern deformed by the object. Therefore, this method can provide a 3D map of objects with pure color.

Besides that, the methods using a grid pattern with the intersection of the slits chosen as feature points have the great advantage of fast decoding and high accuracy in the 3D map. In [5], a De Bruijn sequence-based grid pattern is proposed. The pattern is coded based on the De Bruijn space with horizontal blue slits and red vertical slits. This method and the one using a binary grid pattern introduced in [6] use the spatial constraints to detect the feature points. In [7, 8], a color grid based on the De Bruijn sequence coded in both axes is used. With the different color channels used to encode the vertical and horizontal slit respectively, feature points are precisely detected in the skeleton image. The entire pattern is then fully decoded using the method in [8].

Recently, with the ability of learning complex hierarchies of features from input data, convolutional neural networks (CNNs) are applied to several fields of machine vision. Zhang in [9] succeeds in extracting the line areas from the aerial image with a Deep Residual U-Net with a very small number of parameters compared to the conventional U-net networks. Wang et al. [10] proposed a new framework for segmenting the retinal vessel based on the patch-based learning strategy and dense U-net that works well with the STARE and DRIVE datasets. In 3D reconstruction, Nguyen in [11] proposed an FPP-based single-shot 3D shape reconstruction system and conducted experiments with three different network structures: Fully convolutional networks (FCN), Autoencoder networks (AEN), and UNet. This method can obtain a 3D depth map from its corresponding 2D image by a transformation without any extra processing. However, this method requires a large amount of high-quality 3D ground-truth labels obtained by a multi-frequency fringe projection profilometry technique. For improving the performance of the 3D one-shot imaging system, many CNN-based methods using a binary pattern were introduced. Tang et al. [12] proposed a pattern encoded in the 2x2 property window with eight geometries. Elements of the deformed pattern after extracted from the image are classified with a pre-trained CNN network. The feature point is defined as the intersection of two adjacent rhombic shapes in the property window. Furthermore, Song et al. [13] proposed a binary grid pattern embed with eight geometric shapes. Where the feature points, which are the grid intersections, are detected by applying a cross template over the enhanced image. After extracting based on the four feature points at the four corners around themselves, the elements of the pattern then are classified by a pre-trained CNN model

In the 3D reconstruction methods mentioned above, most results are evaluated only on the density of the point cloud, while a few methods regard the accuracy of the corresponding points reconstructed in the 3D map. To avoid false detection of feature points in areas with extreme distortion, Ha et al. [8] proposed a method for detecting opened-grid-points. However, locating the position of feature points refers to the center of the opened-grid-points at such regions just achieves an acceptable accuracy. In [14], feature points at the objects' boundary regions detected with a large deviation from the groundtruth that reduce the average accuracy of the reconstructed 3D map. Furukawa and Kawasaki et al. [15] proposed a method for detecting the feature points with a pre-trained CNN model. This process is divided into two phases: detecting horizontal and vertical lines individually and detecting feature points. However, the datasets used for training in both phases were manually annotated; therefore, this method is unable to achieve a high location accuracy.

To improve the location accuracy of detected feature points, in this paper we propose a novel method of detecting the intersection points in a grid pattern based on the U-net introduced in our previous work [16]. Firstly, different objects at each pose captured with one shot and two shot imaging methods, respectively. In the first method, a grid pattern projected onto the scene, while the other one uses a pattern encoded in vertical and horizontal axis individually. After applying a morphological skeletonization algorithm, the pictures taken by the two-shot method will be fused in the complete skeleton images and then selected as the groundtruth images. These images and their corresponding gray images were taken by the one-shot method are sequentially cut into patches with the size of 64 x 64 pixels and fed to the training set, whereas, the remaining original large images are the test set.

The remains of this paper are organized as follows. Section 2 gives an overview of the imaging system used in the article. Then, Section 3 produces the detail of the proposed method such as U-net network architecture, data set designing, and hyperparameters setting to train the model. The experimental results and evaluation are presented in Section 4. Finally, the conclusions and future scope of the paper are given in Section 5.

## 2. SYSTEM OVERVIEW

As shown in Figure 1, our system includes two main parts which are an image acquiring subsystem (IAS) and a data processing unit. In which, the IAS consists of a Daheng MER-503-36U3C with a resolution of 2448 x 2048 pixels, a Sony IMX264 sensor, a CMOS global shutter, 36 fps, and a Canon REALiS SX7 LCoS projector with a resolution of 1400 x 1050 pixels. Meanwhile, the data processing unit is a desktop computer with an Intel Core i7-8700, 16GB RAM, and an NVIDIA GeForce RTX 2070 graphics card. Also, two patterns based on De Bruijn sequence coded in the vertical and horizontal direction with 127 slits and 66 slits respectively that are utilized for the two-shots imaging method. And a color grid pattern combined from those is for the one-shot imaging method [8].

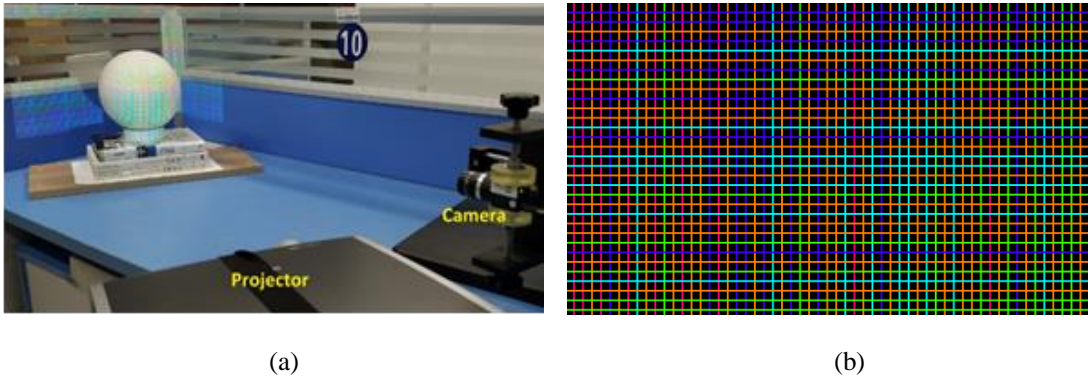

(a)                                                            (b)

Figure 1. Our experimental setup and a part of the color grid pattern are given in (a) and (b), respectively.

## 3. METHODOLOGY

### 3.1. U-Net Architecture

For traditional U-Net networks, since the output data size is smaller than the input one. The pixels near the border of the training image are missing which leads to lost border information.

To avoid that, in our network, padding is applied to ensure that the input and output sizes are the same (as shown in Figure 2). This operation is carried out as follows:

1) Padding of the 0 value around the feature map of W x H is performed.

2) The size of the output feature map is increased by employing the convolution with a filter size of 2 x 2.

After a few layers of convolution and upsampling, the size of the segmented image can be expanded so that it is equal to the size of the input image.

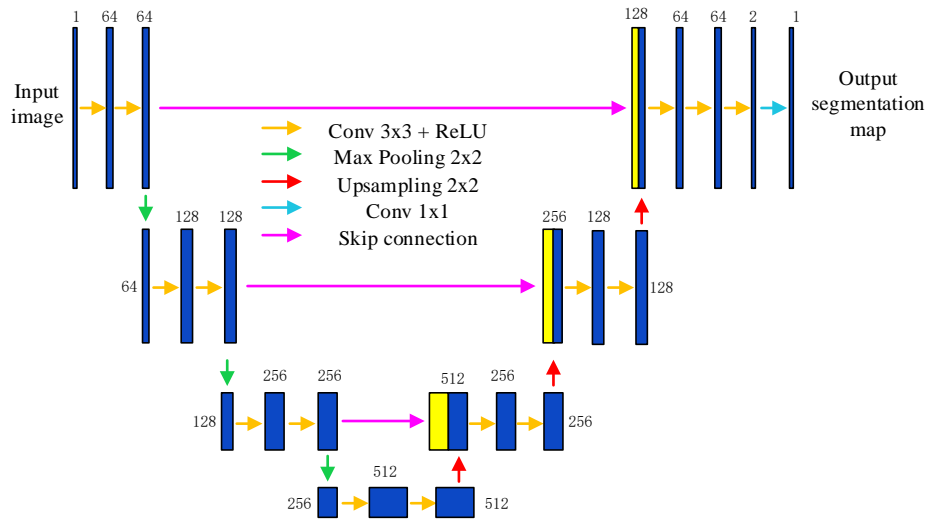

Figure 2. The proposed U-net architecture.

The structure of our CNN consists of a contracting path (left side) and an expansive path (right side), where Conv represents the convolutional layer, Max pooling indicates the maximum pooling layer, Upsampling is the deconvolution layer, and Skip connection operations are carried out by Merge layers. It is worth noting that there are only two categories, including the background and grid, so several hidden layers based on the U-net model are reduced. As a result, there are only a total of 16 convolutional layers and three pools, as well as three upsampling layers and three skip connection structures left. On the contracting path, the size of the input image is gradually reduced by performing six convolutional layers with filters of 3 x 3 pixels with ReLU activation functions and three Max pooling operations with a filter size of 2 x 2. Also, some dropout layers are added to the downsampling section to prevent overfitting. Meanwhile, by using upsampling operations instead of max pooling ones, a similar structure is applied on the expansive path. Thereby, the size of the output image is guaranteed to be the same as the size of the input one.

## 3.2. Data Labeling Method

The analysis of the images obtained from the IAS shows that, in the ideal case, the grid pattern deformed is the bright area, whereas the background is the dark area. However, these assumptions might be broken since the properties of the scene. For example, with fabric or human skin, the light is dispersed, so the light stripes in the captured image are wider, in contrast to the dark stripe. Meanwhile, plaster material with low internal reflectivity results in the image of a sharp and uniform deformed pattern. To minimize the influences of ambient light and color noise, and obtain the best data set for training, a two-shot imaging method was performed to get the label images.

As shown in Figure 3, the label set designed as follows. Firstly, objects at each pose are projected with two patterns based on De Bruijn sequence coded in the horizontal and vertical axis, respectively. The images of the deformed pattern after captured by the camera then are applied some preprocess techniques such as noise-reducing, blurring, and converting to HSV color space then take the value channel as grayscale images. After that, the skeleton images, composed of horizontal and vertical lines respectively, can be obtained by applying a morphological skeletonization algorithm to them. Finally, these images are fused and then sequentially cut into small patches with the size of 64x64 pixels as labeled images.

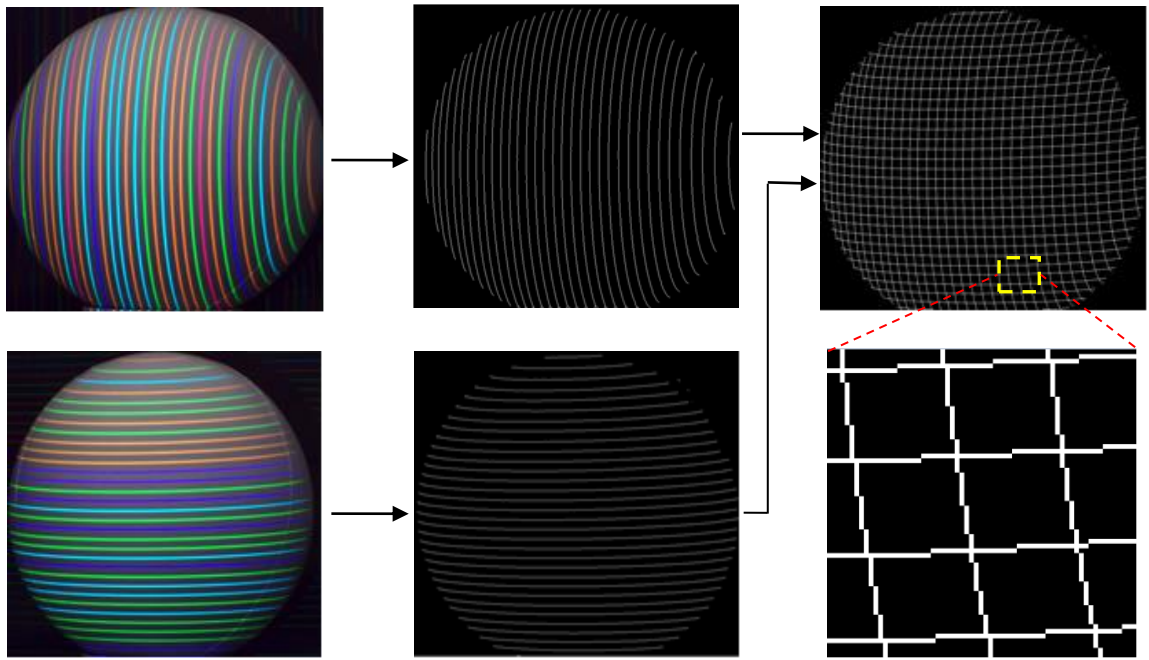

Figure 3. Illustration of the data labeling process.

## 3.3. Data Augmentation

Our dataset consists of 20 original large grayscale images cropped to a size of 1664 x 1664 pixels to conventional for training and testing. Among them, 13 images and their labeled ones are sequentially cut into patches with a size of 64 x 64 pixels. So that, 676 patches are obtained from each large image. 20 % of these patches are chosen as a validation set, and the remaining ones are the training set (Figure 4). By the data augmentation, the training set was expanded, and a new data set of over five million training images was obtained.

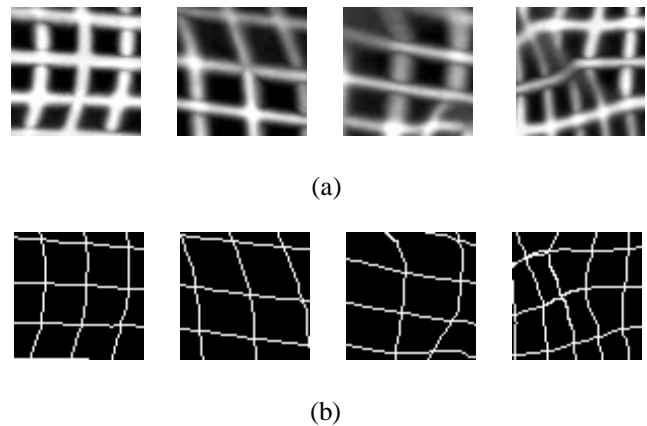

(a)

(b)

Figure 4. Illustration of the image patches of the training set. (a) Original image patches.
(b) Labeled image patches.

## 3.4. Parameter Settings

Before putting the model into training, it is very essential to set the values of hyperparameters as it has a significant influence on the training results. Among them, the learning rate directly affects the learning speed of the model by regulating the intensity of the model's weighting update. If this parameter is set too small, the learning time is extended too long, on the contrary, a too-large value can lead to an unstable learning process. In this study, by setting the learning rate to a value of 1e-3, the network converges quickly at the point closest to the optimum one and the model can learn the detailed features in the dataset.

The batch size indicates the number of training samples utilized in one iteration. The value of this parameter plays an important role in the convergence process of the network. A large batch size might lead the network to fall into the local minimum point and cannot converge at its optimum point. Whereas, the network can converge at the optimal point, but this extends the training time. The batch size used in our network is set to 128. Since an epoch consists of one full cycle through the training data, so aiming at a good training result without costing much time, the model is trained with 100 epochs. In each epoch, the number of steps per epoch, which indicates the number of iterations, is set to 1000. Thus, input samples are randomly selected from the above-designed training set with the batch size for processing at each iteration.

## 3.5. Training and Testing Processes

During training, the samples are continuously parked with the batch size and fed to the network from the augmented training set, where each of them consists of the sample itself and a corresponding label image. For each filter in a layer, the network extracts several features of the input sample and transmits them to the output layer through the intermediate hidden layers. Along with that, the value of all the elements of the weight matrix between the layers is adjusted continuously. After training with several epochs, the value of the loss function is getting smaller and smaller and simultaneously the accuracy also gradually attains the desired value.

A block diagram of the training and testing progress for grid pattern segmentation is shown in Figure 5. Where the input samples are the patches with a size of 64 x 64 pixels drawn from the training set and fed to the proposed CNN. The network is currently trained with such samples and the preset hyperparameters. In each training cycle, the loss function and the accuracy are utilized to monitor and adjust the value of the parameters for the next cycle. Finally, a high-precision segmentation model for accurately segmenting grid pattern is obtained.

After training, the original large gray images taken from the test set are utilized to evaluate the performance of the model. Because it can ensure the size of the input image is the same as the size of the output image, and different sizes of the input will only lead to different sizes of the output, so it was possible to directly predict a test image with a size of 1664 x 1664 pixels.

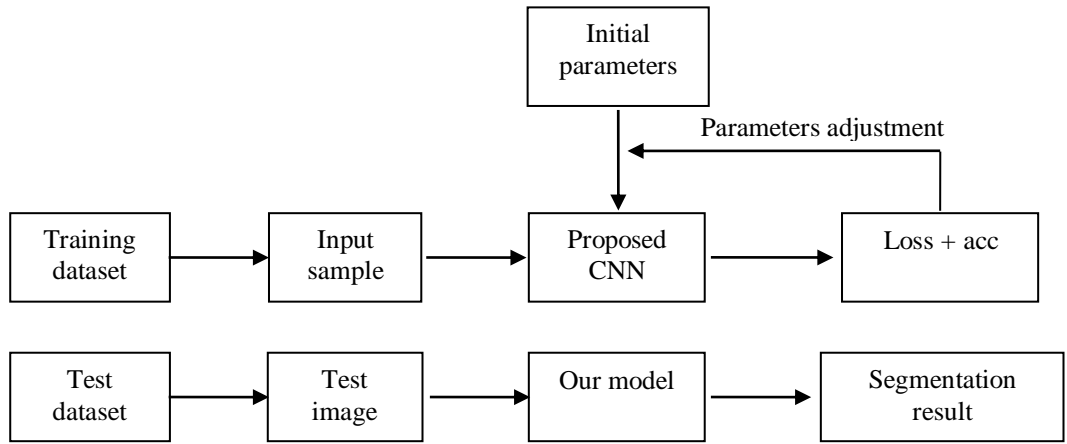

Figure 5. A block diagram of the training and testing process for grid pattern segmentation.

## 3.6. Grid-Point Detection Method based on U-Net

As we all know, the one-shot imaging method first projects an encoded pattern onto the scene. Then the deformed pattern on the surface of objects in the camera image is extracted and sent to a decoding procedure. From the correspondences, those are in the camera image plane and the projector image plane, and the parameters of a calibrated system, a triangulation method is applied to obtain the 3D map of the scene. However, the objects' surface is always inhomogeneous, resulting in very large deformations of the projected pattern. Therefore, it is necessary to have the methods of precisely locating the coordinates of the feature points in the images for an improvement of the quality of 3D point cloud reconstruction.

For the one-shot imaging methods using a grid pattern, in particular, the selected grid intersections are the feature points. Since the inhomogeneous property of the scene, the projected pattern might be extremely contracted or stretched, making it difficult to accurately locate the feature points in the captured images. Huang's method in [9] only detects the feature points in the large and quite planar regions of the objects' surface as consequences of regardless of the ones at the boundaries or extremely deformed. Ha's method in [8] is capable of overcoming the shortcomings of Huang's method by taking the advantages of opened-grid-points, thus obtaining a denser 3D map with higher accuracy. However, this method is still challenged by detecting the grid-points in case of the gap between two consecutive stripes becomes smaller than the width of the stripes themselves, and the detection accuracy needs to be improved.

With the advantages of CNN in the field of image segmentation and classification, the method for grid-point detection proposed in this article is shown in Figure 6. Firstly, the image captured by the one-shot and two-shot methods after preprocessed are assigned to the training set and test set, individually. After training the network with the designed architecture and initial parameters, the model for grid-point segmentation can be obtained. With this model, the skeleton image of the input grayscale image is quickly obtained. Therefore, it is possible to provide a method for detecting grid-points with higher accuracy.

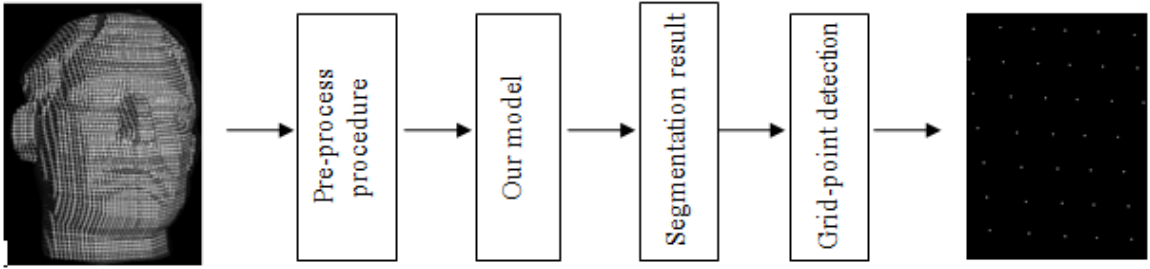

Figure 6. Flow chart of the proposed grid-point detection method.

## 4. EXPERIMENTAL RESULTS AND DISCUSSIONS

### 4.1. Quantitative Evaluation of the Grid-Point Detection Method

To estimate the flexibility of the proposed method, the test dataset includes seven images of different objects, i.e, a planar board, a rotating fan blade, a geometrical shape, plaster models, and a sphere. With the difference in imaging conditions such as the different ambient light and the dynamic scene, the comparative experiment results are shown in Figure 7.

As shown in Table 1, regardless of the grid-points that were extremely deformed at the boundary regions of the objects, the method of Huang et al. [9] can only detect the ones in the large regions of the object's surface. Overcome this shortcoming, the method introduced by Ha et al. [8] detected most intersections of the grid pattern on the object's surface with the benefits of the opened-grid-point detection method. Even better than them, our method can segment the grid-points that were broken, blurred, or adhesive especially objects with complex surfaces, such as the human hand and the plaster models.

For objects with a relatively homogeneous surface such as a planar board, all three methods detect most of the feature points in the image. For the other subjects, Ha's method detected more feature points than Huang's method with the ones strongly stretched and in the boundary regions. However, in case the distance between horizontal or longitudinal stripes is smaller than the width of a stripe, the feature points there might be adhesive and leading to false detections. With the proposed method, the stripes of the deformed grid in the image obtained with our model are thinned out to a width of 3 pixels, while increasing the distance of the adjacent stripes, thereby solving the abovementioned problems. Along with that, the strongly deformed intersections in the original image are narrowed down in the width of the stripes in the resulting image, so the size of the opened-grid-points in the final feature detection image at the corresponding positions is considerably smaller compared to the ones obtained with Ha's method. As a result, the appearance of the too-large opened-grid-points is no longer exist. With the above advantages, the feature points can be detected more easily with our method.

Table 1. Grid-point detection results of different objects with different methods.

| Object | Planar board (pixels) | Fan blade (pixels) | Geometrical shape (pixels) | Plaster model (pixels) | Standard sphere (pixels) |
|---|---|---|---|---|---|
| Huang's method | 2497 | 476 | 1466 | 1325 | 782 |
| Ha's method | 2509 | 561 | 1603 | 1477 | 869 |
| Our method | **2523** | **568** | **1611** | **1480** | **873** |
| Groundtruth | 2537 | 572 | 1619 | 1482 | 874 |

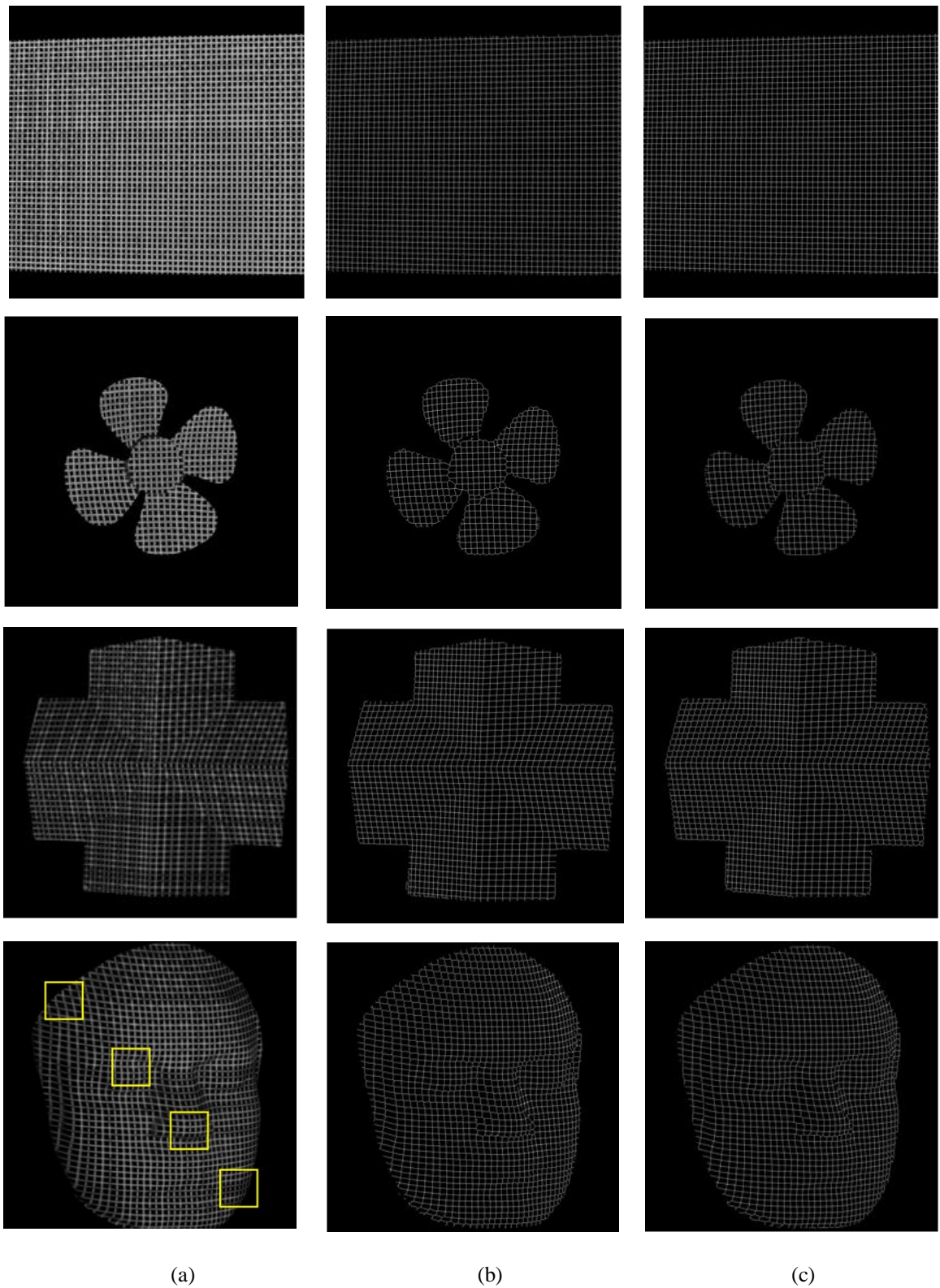

(a) (b) (c)

Figure 7. Segmentation results with different methods. (a) Original gray images. (b) Ha's method in [8], (c) Our method

## 4.2. Evaluation of Detection Accuracy

With the patches marked as yellow squares in Figure 7, Figure 8 illustrates a comparison of the segmentation result of a plaster model with Ha's method in [8] and ours. It can be seen that the opened-grid-points with large size detected by Ha's approach extremely distort the obtained image and makes the horizontal and vertical lines less smooth than the result obtained by our method. Furthermore, such large opened-grid-points surely impact the location accuracy of the feature points detected. Moreover, it is necessary to remove some virtual segments that appeared in the resulting image of the previous method before detecting feature points.

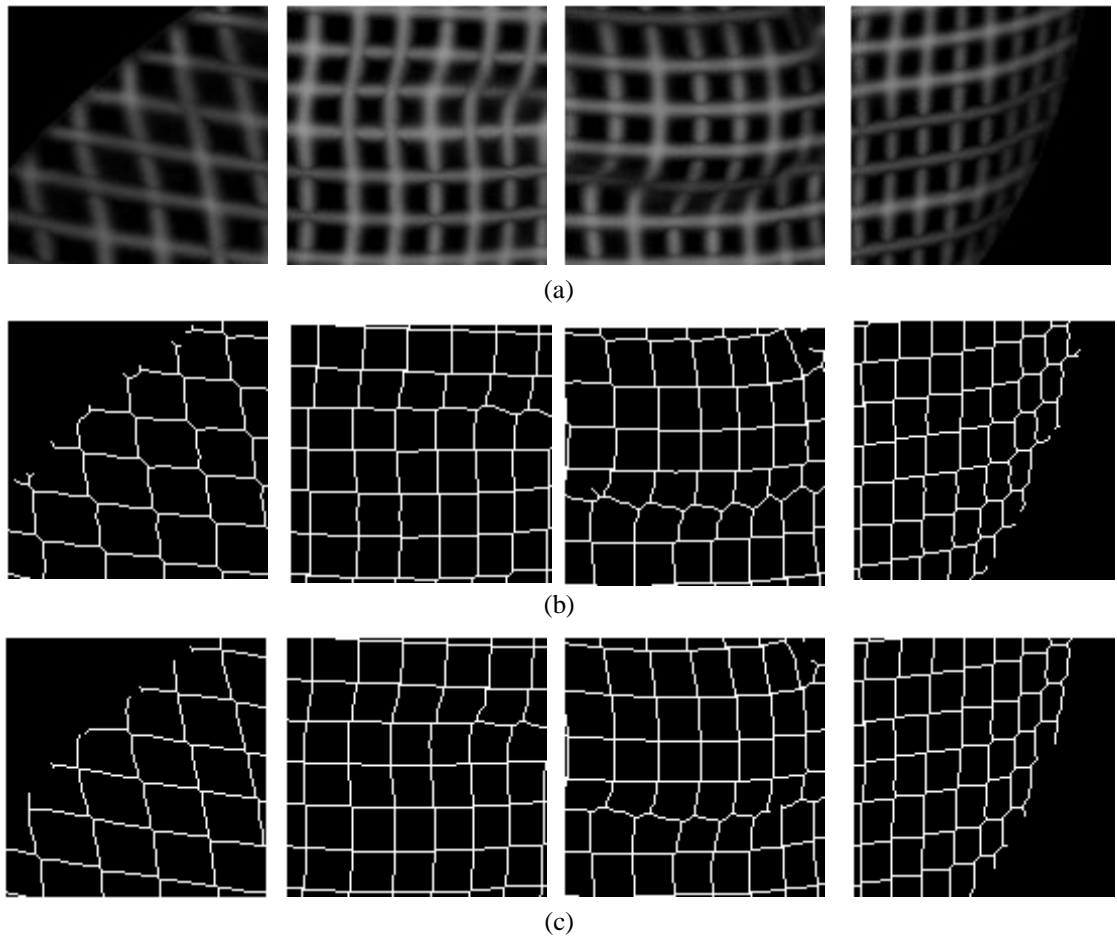

Figure 8. Some patches of segmentation result image of a plaster model with two methods. (a) Original gray images. (b) Ha's method in [8], (c) Our method

To further evaluate the detection accuracy of the proposed method, an experiment was conducted with a standard sphere. Along with that, a comparison of the proposed method and the one introduced in [8] is carried out. A groundtruth map of intersections is obtained by performing the two-shot imaging method. The experimental results are shown in Figure 9. In which the blue points indicate the grid-points detected with the two methods, and the red ones are the true locations of the grid intersections. In which the positions with only one red point indicate the positions of two points that are considered to be completely matched. It can be seen from Figure 9, the deviation of the feature points detected by our method to the ones in the groundtruth image is smaller than the ones with Ha's method. Furthermore, the grid intersections in Figure 9 are stretched relatively large so these deviations might be greater than 1 pixel, but they are approximately zero in the quite flat regions.

For a pixel location accuracy of the detected feature points, we define a mean absolute error (MAE) calculated by the formula (1):

$$MAE = \frac{\sum_{t=1}^{N} |C_t - C_d|}{N}, \qquad (1)$$

Where N denotes the total of detected feature points, Cd is the coordinate of the ones detected in the final image obtained with the two methods, and Ct represents the coordinate of the ones in the groundtruth image.

It worth noting that, with the feature points are the centers of the opened-grid-points, the pixel location accuracy of our method is higher than the other proposed in [8] with a MAE value of 0.27 pixels and a maximum error of 2 pixels, while those of Ha's method are 0.33 and 2 pixels respectively.

In addition, another index here defined to evaluate the local accuracy of the feature points detected by the above two methods that calculated by the formula (2):

$$D = \frac{N_1}{N} \cdot 100\%, \qquad (2)$$

Where $N_1$ denotes the quantitive of detected feature points with a deviation of less than 1 pixels to the ones in the groundtruth image, and N indicates the quantitive of all of the detected ones. The experimental results show that this index achieved with our method is 91% which is better than 78% with Ha's method.

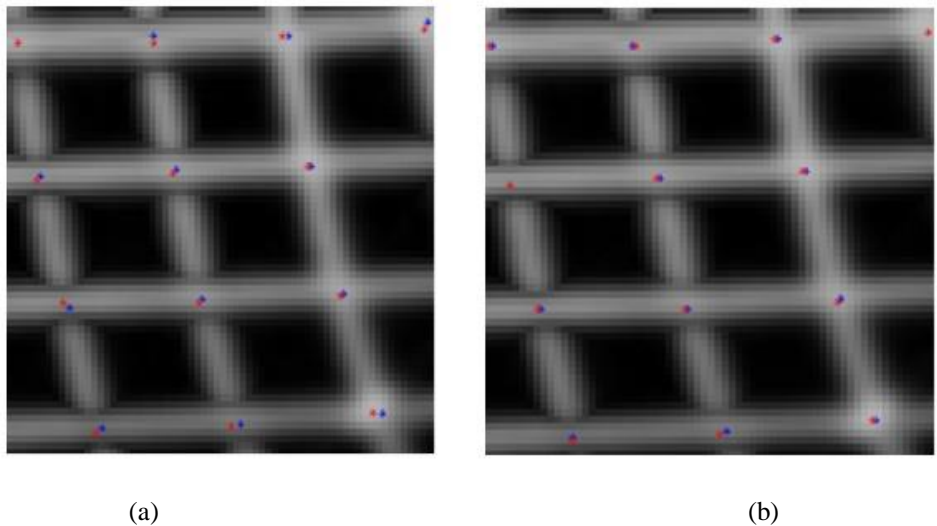

(a)          (b)

Figure 9. Illustration of a pixel location accuracy comparison between the feature points detected with the method in [8] (a) and the proposed method (b) (marked in blue) and the ones in the groundtrurth image (marked in red).

Besides the good results obtained during testing and evaluation of the proposed model, our method still has some following limitations:

1) Although using a training set with the labels composed of binary single lines, the model still produces a grid image with 3 pixels wide stripes, especially for some strongly distorted areas where the stripes' width might reach 4 pixels. Hence in the final result image, there still exist some opened-grid-points like Ha's method, however, their size at the corresponding locations has been significantly reduced.

2) It is quite hard to detect the feature points in a completely new image of a dynamic scene if there are no images of similar objects acquired with the two-shot imaging method in the training set.

## 5. CONCLUSIONS

This paper has proposed an approach for detecting the grid-points of the grid pattern for a structured light system mainly based on a U-net network. Along with that, a specific dataset is designed for training and testing the model. Whereas, the training set includes the patches of original large gray images and their labeled images, which are captured by utilizing the two-shot imaging method. And the test set consists of grayscale images that are completely different from the ones in the training set. A comparison between the experimental results obtained with our method and the two others shows that our method provides better performance. Furthermore, to evaluate the pixel location accuracy of the feature points detected, an experiment with a standard sphere is conducted. The experimental result proves that the proposed method can achieve a higher location accuracy comparing with Ha's in [8]. On the other hand, our method is convenient to apply to reality applications since it is compatible with both color and monochromatic grid patterns. Moreover, the quality of the point cloud reconstructed can be improved with such a high performance of feature points detected. To improve the proposed method, collecting more images of different objects, materials, with different imaging conditions to enrich the training set, and also eliminating the opened-grid-points in the final image for higher location accuracy of detected feature points is our future scope.

## REFERENCES

[1] Joaquim Salvi, Sergio Fernandez, Tomislav Pribanic, and Xavier Llado, (2010) "A state of the art in structured light patterns for surface profilometry." Pattern Recognition Vol. 43, No. 8, pp 2666-2680.
[2] Xu Zhang, Youfu Li and Limin Zhu, (2012) "Color code identification in coded structured light." Applied Optics Vol. 51, No. 22, pp 5340-5356.
[3] Yang Lei, Kurt R. Bengtson, Lisa Li, Jan P. Allebach, (2013) "Design and decoding of an M-array pattern for low-cost structured light 3D reconstruction systems." 2013 IEEE International Conference on Image Processing, pp 2168-2172.
[4] Tomislav Petković, Tomislav Pribanić and Matea Đonlić, (2016) "Single-shot dense 3D reconstruction using self-equalizing De Bruijn sequence." IEEE Transactions on Image Processing Vol. 25, No. 11, pp 5131-5144.
[5] Ali Osman Ulusoy, Fatih Calakli and Gabriel Taubin, (2009) "One-shot scanning using De Bruijn spaced grids." IEEE 12th International Conference on Computer Vision Workshops, pp 1786-1792.
[6] Guangming Shi, Ruodai Li, Fu Li, Yi Niu and Lili Yang, (2018) "Depth sensing with coding-free pattern based on topological constraint." Journal of Visual Communication and Image Representation Vol. 55, 229-242.
[7] Bingyao Huang and Ying Tang, (2014) "Fast 3D reconstruction using one-shot spatial structured light." 2014 IEEE International Conference on Systems, Man, and Cybernetics, pp 531-536.

Computer Science & Information Technology (CS & IT) 39

[8] Minhtuan Ha, Changyan Xiao, Dieuthuy Pham and Junhui Ge, (2020) "Complete grid pattern decoding method for a one-shot structured light system." Applied Optics, Vol. 59, No. 9, pp 2674-2685.
[9] Zhengxin Zhang, Qingjie Liu and Yunhong Wang, (2018) "Road extraction by deep residual u-net." IEEE Geoscience and Remote Sensing Letters Vol. 15, No. 5, pp 749-753.
[10] Chang Wang, Zongya Zhao, Qiongqiong Ren, Yongtao Xu and Yi Yu, (2019) "Dense u-net based on patch-based learning for retinal vessel segmentation." Entropy Vol. 21, No. 2, pp 168.
[11] Hieu Nguyen, Yuzeng Wang and Zhaoyang Wang, (2020) "Single-Shot 3D Shape Reconstruction Using Structured Light and Deep Convolutional Neural Networks." Sensors Vol. 20, No. 13, pp 3718.
[12] Suming Tang, Xu Zhang, Zhan Song, Hualie Jiang and Lei Nie, (2017). "Three-dimensional surface reconstruction via a robust binary shape-coded structured light method." Optical Engineering Vol. 56, No. 1, 014102.
[13] Zhan Song, Suming Tang, Feifei Gu, Chua Shi and Jianyang Feng, (2019) "DOE-based structured-light method for accurate 3D sensing." Optics and Lasers in Engineering, Vol. 120, 21-30.
[14] Suming Tang, Xu Zhang, Zhan Song, Lifang Song and Hai Zeng, (2017) "Robust pattern decoding in shape-coded structured light." Optics and Lasers in Engineering Vol. 96, pp 50-62.
[15] Ryo Furukawa, Daisuke Miyazaki, Masashi Baba, Shinsaku Hiura and Hiroshi Kawasaki (2019), "Robust structured light system against subsurface scattering effects achieved by CNN-based pattern detection and decoding algorithm." ECCV Workshop 3D Reconstruction in the Wild, pp 372-386.
[16] Dieuthuy Pham, Minhtuan Ha, San Cao and Changyan Xiao, (2020) "Accurate stacked-sheet counting method based on deep learning." Journal of the Optical Society of America A, Vol. 37, No. 7, pp 1206-1218.

## AUTHORS

**Minhtuan Ha** received the B.S degree in automation from Viet Nam Maritime University, Hai Phong, Viet Nam, in 2005, and the M.S. degree in Measurement and Control systems from Ha Noi University of Science and Technology, Ha Noi, Viet Nam, in 2010. He is currently pursuing a Ph.D. degree with the College of Electrical and Information Engineering, Hunan University, Changsha, China. His current research interests include structured light systems, 3D imaging, machine vision and machine learning.

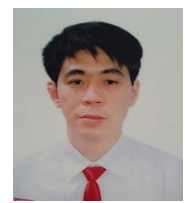

**Dieuthuy Pham** received the B.S. degree in automation from Thai Nguyen University, College of Engineering, Thai Nguyen, Viet Nam, in 2006 and the M.S. degree in Measurement and Control systems from Ha Noi University of Science and Technology, Ha Noi, Viet Nam, in 2010. She is currently pursuing a Ph.D. degree with the College of Electrical and Information Engineering, Hunan University, Changsha, China. Her current research interests include medical image processing, machine vision and machine learning.

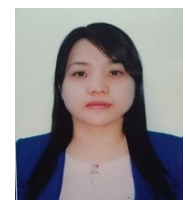

**Changyan Xiao** received the B.E. and M.S. degrees in mechanical and electronic engineering from the National University of Defense Technology, Changsha, China, in 1994 and 1997, respectively, and the Ph.D. degree in biomedical engineering from Shanghai Jiaotong University, Shanghai, China, in 2005., From 2008 to 2009, he was a Visiting Postdoctoral Researcher with the Division of Image Processing, Leiden University Medical Center, Leiden, Netherlands. Since 2005, he has been an Associate Professor and a Full Professor with the College of Electrical and Information Engineering, Hunan University, Changsha. His current research interests include medical imaging and machine vision.

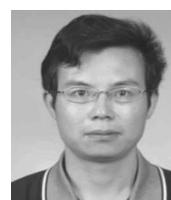